\documentclass[a4paper,fleqn]{cas-sc}
\usepackage[authoryear,longnamesfirst]{natbib}
\usepackage{graphicx}
\usepackage{booktabs}
\usepackage{multirow}
\usepackage{array}
\usepackage{longtable}
\usepackage{tabularx}
\usepackage{amsmath}
\usepackage{amssymb}
\usepackage{url}
\usepackage{hyperref}
\usepackage{float}
\usepackage{threeparttable}
\usepackage{pdflscape}
\usepackage{caption}
\usepackage{algorithm}
\usepackage{algpseudocode}

\def\tsc#1{\csdef{#1}{\textsc{\lowercase{#1}}\xspace}}
\tsc{WGM}
\tsc{QE}

\begin{document}
	\renewcommand{\floatpagefraction}{.7}
	\renewcommand{\textfraction}{.15}
	\renewcommand{\topfraction}{.85}
	\renewcommand{\bottomfraction}{.65}
	\setcounter{topnumber}{2}
	\setcounter{bottomnumber}{1}
	\setcounter{totalnumber}{3}

	\shorttitle{Seq103: A Unified Neuroevolution Framework for Compact Sequence Architecture Discovery}
	\shortauthors{Li et al.}
	
	\title[mode=title]{Seq103: A Unified Neuroevolution Framework for Compact Sequence Architecture Discovery}
	\author[1]{Wenxiao Li}
	\ead{321145@whut.edu.cn}
	\credit{Conceptualization, Methodology, Software, Validation, Formal analysis, Investigation, Writing -- original draft}
	
	\author[1]{Yongjian Liu}
	\credit{Supervision, Writing – review and editing, Funding acquisition}
	
	\author[1]{Qing Xie}\cormark[1]
	\ead{felixxq@whut.edu.cn}
	\credit{Supervision, Writing – review and editing, Project administration}

	\affiliation[1]{organization={School of Computer Science and Artificial Intelligence, Wuhan University of Technology},
		addressline={No. 122 Luoshi Road, Hongshan District},
		city={Wuhan},
		postcode={430070},
		state={Hubei},
		country={China}}
	
	\cortext[1]{Corresponding author.}
	
	\begin{abstract}
		Neuroevolution is a representative neural architecture search paradigm that evolves both network topology and weights through evolutionary algorithms. In this paper, we propose Seq103, a unified NEAT-style neuroevolution framework for compact sequence architecture discovery. Seq103 consists of a shared evolutionary backbone and an optional recurrent extension. The shared backbone includes an elementary node-and-connection representation, per-class RMSE-based evaluation, mutation-based evolution with class-wise recombination, and elitism. The optional hidden-state mechanism extends the search space with hidden-state nodes and hidden connections, enabling temporal memory when step-wise recurrent inference is required. With this design, Seq103 applies the same core search pipeline to both step-wise recurrent and sample-wise feedforward sequence classification. In recurrent tasks, the hidden-state extension is enabled to provide temporal memory; in feedforward tasks, it is disabled while the shared evolutionary backbone remains unchanged. We evaluate Seq103 on 8 text classification datasets and the full UCRArchive2018 benchmark with 128 univariate time-series datasets. On step-wise tasks, Seq103 retains 86.96\% of the best-baseline accuracy on average while using 34.6$\times$ to 3218.0$\times$ fewer parameters. On sample-wise tasks over the full UCRArchive2018 benchmark, Seq103 retains 81.95\% of the best-baseline accuracy on average while using 11.8$\times$ to 160,601.0$\times$ fewer parameters.
		
	\end{abstract}
	
	\begin{highlights}
		\item A NEAT-style elementary search space extended with optional hidden-state mechanisms for recurrent sequence modeling.
		\item A shared neuroevolution framework that handles both step-wise recurrent and sample-wise feedforward classification.
		\item Competitive performance with 34.6$\times$ to 3218.0$\times$ fewer parameters on step-wise tasks and 11.8$\times$ to 160,601.0$\times$ fewer parameters on sample-wise tasks.
	\end{highlights}

	
	\maketitle
	
	\section{Introduction}
	
	Sequence classification is a fundamental problem in many application domains, including text understanding, time-series analysis, speech processing, and sensor signal recognition. In practice, two common processing regimes are widely used: step-wise recurrent processing, where a model consumes an input sequence one step at a time and updates an internal state over time; and sample-wise feedforward processing, where the entire sequence is represented as a single sample and processed in one forward pass. Although both belong to sequence classification, they rely on very different information-flow patterns and are usually modeled with different architectures. As a result, effective model design often depends heavily on expert knowledge and prior experience.
	
	Neural architecture search (NAS) methods can reduce this manual effort to some extent by formulating architecture design as an optimization problem. A typical NAS pipeline contains three components: the search space, the search strategy, and the evaluation module. In this area, neuroevolution is a representative approach that uses evolutionary algorithms to optimize both topology and weights.
	
	To further reduce reliance on expert knowledge and task-specific priors, we focus on two design goals for a unified neuroevolution framework for sequence classification. First, the search space should rely as little as possible on predefined architectural modules such as blocks or cells, so that topology can be explored at a more elementary level. Second, the framework should support different information-flow patterns, including step-wise recurrent processing and sample-wise feedforward processing, under a shared search pipeline.

	In this paper, we propose Seq103, a unified NEAT-style neuroevolution framework for sequence architecture discovery. Seq103 is built around a shared evolutionary backbone and an optional recurrent extension. The shared backbone includes an elementary node-and-connection representation, per-class RMSE-based evaluation, mutation-based evolution with class-wise recombination, and elitism. These mechanisms are used in both step-wise recurrent and sample-wise feedforward settings. The optional hidden-state mechanism is a search-space capability that is enabled only when temporal memory is required. It introduces hidden-state nodes and hidden connections, allowing archived node states to be propagated across time steps. In step-wise recurrent tasks, the hidden-state mechanism works together with the shared evolutionary backbone to evolve recurrent sequence models. In sample-wise feedforward tasks, the hidden-state mechanism is disabled, while the same evolutionary backbone remains unchanged. Thus, Seq103 handles both processing regimes within one framework rather than relying on two separate task-specific methods.
	
	We evaluate Seq103 on two representative benchmark settings: 8 text classification datasets for the step-wise recurrent regime and all 128 datasets in UCRArchive2018 for the sample-wise feedforward regime. Experimental results show that Seq103 can discover classifiers with competitive performance and compact architectures across both processing regimes. On step-wise tasks, Seq103 achieves 73.6\% to 99.4\% of the accuracy of the best baselines with 34.6 $\times$ to 3218.0 $\times$ fewer parameters. On sample-wise tasks, it achieves 5.2\% to 117.3\% of the accuracy of the best baselines with 11.8 $\times$ to 160,601.0 $\times$ fewer parameters.
	
	The main contributions of this work are as follows:
	\begin{itemize}
		\item We extend a NEAT-style elementary neuroevolution search space with optional hidden-state mechanisms, enabling the representation and evolution of recurrent sequence architectures.
		\item Seq103 supports both step-wise recurrent inference and sample-wise feedforward inference under a unified evolutionary framework, while activating or deactivating the hidden-state mechanism according to the target information-flow pattern.
		\item We demonstrate that the discovered architectures achieve competitive performance while remaining highly compact on text classification and UCR time-series classification benchmarks.

	\end{itemize}
	
	\section{Related Work}
	
	In this section, we first introduce the two representative processing regimes in sequential classification, together with the corresponding tasks and baselines. We then briefly review NAS methods, especially neuroevolution methods.
	
	\subsection{Tasks and baselines}
	
	Sequential classification problems contain two representative processing regimes: (i) step-wise recurrent processing, where models process inputs one step at a time and make predictions from the final-step output; and (ii) sample-wise feedforward processing, where the entire sequence is mapped to a label in one forward pass. Although both are sequence classification settings, they differ significantly in information-flow patterns and are usually associated with different modeling choices.
	
	\subsubsection{Step-wise recurrent sequence classification}
	
	In this formulation, each example is a variable-length sequence, with a label that can be either multi-class or multi-label. Models process the sequence sequentially, update states over time, and output a single prediction from the final-step output.
	
	We instantiate this regime with 8 text classification benchmarks, including SST-2~\citep{socher-etal-2013-recursive}, IMDB~\citep{maas-EtAl:2011:ACL-HLT2011}, Rotten Tomatoes Movie Review~\citep{Pang+Lee:05a}, and three datasets introduced by \citet{Zhang2015CharacterlevelCN}: AG News, Yelp Review Full, and DBPedia14 for binary and multi-class evaluation. We use Reuters-21578~\citep{APTE94} and GoEmotions~\citep{demszky2020goemotions} for multi-label evaluation.
	 
	We select five baselines as follows:
	\begin{itemize}
		\item \textbf{FastText}~\citep{joulin2016bag}, a lightweight model that represents text by averaging word and n-gram embeddings.
		\item \textbf{TextRNN}~\citep{liu2016recurrent}, a typical recurrent model that updates hidden states recurrently and classifies via the final state.
		\item \textbf{TextCNN}~\citep{kim2014cnn}, a convolutional model for text tasks that applies multiple convolution kernels for feature extraction.
		\item \textbf{HAN}~\citep{yang2016han}, a strong model for long documents with a word-to-sentence hierarchy and attention mechanisms.
		\item \textbf{Transformer} (non-pretrained)~\citep{vaswani2017attention}, built on multi-head self-attention and positional encoding, providing strong parallel sequence modeling.
	\end{itemize}
	
	\subsubsection{Sample-wise feedforward sequence classification}
	
	In this form, each sample is processed as a whole in one forward pass.

	We instantiate this regime using univariate time-series classification benchmarks from the UCR Time Series Classification Archive, which contains 128 datasets across multiple domains, including body gesture signals, industrial sensor signals, images, and other real-world recordings.

	Prior work has explored multiple competitive methods, and we select nine of them as baselines:
	\begin{itemize}
		\item \textbf{MLP}, a simple feedforward baseline.
		\item \textbf{ResNet for TSC}~\citep{wang2017tsc}, which uses residual blocks with skip connections and global pooling.
		\item \textbf{InceptionTime}~\citep{fawaz2020inceptiontime}, which stacks inception modules to extract multi-scale temporal features.
		\item \textbf{TWIESN}~\citep{bagnall2017bakeoff}, which maps timestamps into a reservoir and aggregates per-time-step class probabilities for prediction.
		\item \textbf{FCN}~\citep{wang2017tsc}, a typical fully convolutional network with stacked convolution layers.
		\item Other frequently reported CNN variants, including \textbf{TLeNet}~\citep{leguennec2016tsaugmentation}, \textbf{MCDCNN}~\citep{zheng2014mcdeep}, and \textbf{TimeCNN}~\citep{zhao2017timecnn} (hereafter denoted as \textbf{CNN}).
		\item \textbf{Encoder}~\citep{serra2018encoder}, an attention-based convolutional architecture that replaces global average pooling with an attention mechanism to emphasize informative temporal regions.
	\end{itemize}
	
	\subsection{Neural architecture search and neuroevolution}
	
	Neural architecture search (NAS) transforms experience-based network design into a computational optimization problem and solves it with a specific search strategy within a predefined search space. It can automatically discover architectures that fit the given data. A typical NAS pipeline contains three components: (i) the search space, (ii) the search strategy, and (iii) the evaluation module.
	
	Neuroevolution is a type of NAS method that uses evolutionary algorithms as the search strategy. One representative method is NEAT (NeuroEvolution of Augmenting Topologies)~\citep{stanley2002neat}, which introduces innovation numbers, speciation, and incremental complexification. These mechanisms make NEAT effective for topology search. 
	
	In this work, we build a NEAT-style neuroevolution method, which adopts incremental growth and historical marking mechanisms, and the implementation is based on an open-source NEAT framework~\citep{McIntyre_neat-python}. 
	
	\section{Method}
	
	In this paper, we propose a neuroevolution-based neural architecture search (NAS) framework centered on an extension of a NEAT-style elementary search space with optional hidden-state mechanisms. This extension enables step-wise recurrent inference while preserving compatibility with sample-wise feedforward inference under the same overall search pipeline. The two regimes differ only in the data loader and a small set of hyperparameters.

	To maximize exploration and reduce reliance on predefined functional components such as blocks or cells, we adopt an elementary search space and extend it with a hidden-state mechanism inspired by recurrent neural networks, enabling the model to process information across time steps. The extended search space consists of one type of node with optional hidden-state functionality, and two types of connections: forward connections and hidden connections. As a result, the same search space can represent both feedforward computation and recurrent temporal computation.
		
	To better support compact architecture search, we design a modified evolutionary strategy that uses mutation-based evolution with class-wise recombination. The recombination mechanism assembles new individuals by combining class-specific subgraphs selected from candidate models. This strategy is used in both step-wise recurrent and sample-wise feedforward settings, while hidden connections are enabled only when recurrent temporal memory is required.
	
	\label{sec:method}
	
	\subsection{Search strategy}
	Classical genetic algorithms evolve candidate solutions through initialization, evaluation, and generation of new populations. The most common variation operators are crossover and mutation. In this work, we adopt a NEAT-inspired evolutionary search with three modifications:
	\begin{enumerate}
		\item \textbf{Removing speciation.} We remove the speciation mechanism because we observed that it often reduces the effectiveness of the evolutionary process on our tasks. Instead, we use elitism to preserve top individuals and maintain diversity.
		\item \textbf{Class-wise recombination.} We generate new individuals by selecting stronger class-specific subgraphs and assembling them into offspring.
		\item \textbf{Removing standard crossover.} We remove standard crossover and instead assemble offspring through class-wise recombination followed by mutation.
	\end{enumerate}
	
	As a result, each iteration contains the following steps:
	\begin{enumerate}
		\item evaluating individuals,
		\item computing per-class fitness,
		\item selecting class-specific subgraphs via tournament selection,
		\item assembling offspring by combining the selected subgraphs, and
		\item applying mutation operations to the assembled offspring. Hidden connections are mutated in the same way as forward connections, except that they can only be added between hidden-state nodes.
	\end{enumerate}

	The search procedure is summarized in Algorithm~\ref{alg:search}.
	\begin{algorithm}[t]
		\caption{Seq103 Evolutionary Search Process}
		\label{alg:search}
		\begin{algorithmic}[1]
			\State \textbf{Input:} training data $D$, inference mode $m \in \{\text{step-wise recurrent}, \text{sample-wise feedforward}\}$
			\State Initialize population $P$ with size $N$
			\While{stopping criterion is not met}
			\State Evaluate each individual in $P$
			\State Compute overall fitness and per-class fitness for each individual
			\State Preserve elite individuals $\mathcal{E}$
			\State Initialize offspring set $O$
			\While{$|O| < N - |\mathcal{E}|$}
			\State Initialize a child individual
			\For{each class $c$}
			\State Select a candidate class-specific subgraph by tournament selection based on class-$c$ fitness
			\EndFor
			\State Assemble the child by combining the selected class-specific subgraphs
			\State Apply mutation to the assembled child
			\State Add the mutated child to $O$
			\EndWhile
			\State Set $P \leftarrow \mathcal{E} \cup O$
			\EndWhile
			\State Save the whole population
		\end{algorithmic}
	\end{algorithm}

	\subsection{Search space}
	
	The search space is based on the elementary representation proposed in NEAT and is extended with a hidden-state mechanism inspired by recurrent neural networks.
	
	\subsubsection{Connections}
	
	A connection links nodes and serves as the basic unit of information flow, thereby determining the computation graph of the entire network. Each connection contains: (i) a weight, which is a floating-point trainable parameter and is multiplied by the output of the predecessor node at the current time step; and (ii) an enabled flag, which indicates whether the connection is active. Only active connections participate in the feedforward process. Forward connections can connect both normal nodes and hidden-state nodes, subject to the DAG constraint.
	
	\subsubsection{Hidden connections}
	
	The hidden connection links nodes that are in the \texttt{true} hidden state (hidden nodes), enabling the hidden states flow among hidden nodes, this empowers the network to capture temporal memory, which is crucial for enabling the network to handle time-step recurrent tasks effectively. Each hidden connection contains: (i) a hidden weight, a float value that serves as a trainable parameter and is multiplied by the output of the predecessor hidden node from the previous time step; and (ii) a hidden enabled flag, which indicates whether the hidden connection is active. Only active hidden connections participate in the computation of the hidden state. In this study, all possible topologies of hidden connections are allowed.
		
	\subsubsection{Nodes}
	
	Nodes are elementary computational units, each contains: (i) an archive, which stores the node output from the previous time step, this attribute does not participate in the evolutionary process, and is only used as temporary storage used during evaluation; (ii) a hidden-state flag, if it is \texttt{true}, the node is a hidden-state node, otherwise, it is a normal node; (iii) a bias, which is a regular trainable parameter; (iv) an activation function, with four options for non-output nodes: identity, sigmoid, ReLU, and tanh; (v) an aggregation function, that aggregates the outputs of predecessor normal nodes to produce the necessary input to this node, using one of three operations: sum, mean, or product; and (vi) a hidden-state aggregation function, that aggregates the outputs of predecessor hidden-state nodes to produce the optional input to this node, using one of three operations: sum, mean, or product. 
	In this work, for output nodes, we use task-specific activation functions: identity for single-label classification and sigmoid for multi-label classification.
	
	For a hidden node, the internal computation consists of two parts: the first part aggregates the inputs from the preceding connections, and the second part aggregates the inputs from the preceding hidden connections. These two parts are then summed together, added with a bias, and finally passed through the node's activation function. The resulting value serves as the node's final output, which is propagated backward and used to update the archive. For a normal node, there is only the first part for the final output.
	
	\subsection{Class-wise recombination and evaluation}
	
	We evaluate newly generated individuals directly on the target dataset at two levels.
	
	\subsubsection{Class-level evaluation}
	
	To guide evolution in the correct direction, we define a class subgraph for each output node (class). For each output node, we trace backward through all ancestor nodes and edges that contribute to that output, thereby obtaining a class-specific induced subgraph. Child individuals are then assembled as the union of class-specific subgraphs, with no topology shared across classes (except for input pins).
	
	For model evaluation, we use a class-wise error-based fitness criterion computed from the raw output responses of each candidate network. For each sample, the network produces one output value for every class. Instead of using a conventional one-hot target representation, we adopt a fixed scaled bipolar target scheme, in which the ground-truth class is assigned a positive target value while all remaining classes are assigned a corresponding negative target value. This encourages the candidate network to increase the response of the correct class and suppress the responses of non-target classes.
	
	For each class, the discrepancy between the predicted raw outputs and the corresponding bipolar targets is accumulated over the dataset and summarized using a root-mean-square error criterion. The resulting class-wise error reflects how well the network matches the desired response pattern for that class, and it is used as the basis for computing the candidate network's fitness. Candidate networks with lower class-wise errors are considered to have better predictive behavior and are therefore assigned more favorable fitness values during evolution. The same bipolar target scale is applied consistently across all experiments.

	Based on this fitness, for each class we perform tournament selection across individuals. The winning class subgraph is selected as the corresponding class component of the new individual, and this process is repeated for all classes in all new individuals.
	
	\subsubsection{Individual-level evaluation}
	
	We use overall accuracy to report the final performance.

	\section{Experiments and Results}
	
	In this section, we evaluate our method on both forms of sequential classification tasks. We first introduce the datasets and settings in detail, and then report and analyze the results. Our method is evaluated on a server with two HUAWEI Kunpeng 920 CPUs (32 cores at 2.6 GHz each), for a total of 64 physical cores and 128 threads. All baseline models are evaluated on a server with one AMD EPYC9654 CPU (96 cores at 2.4 GHz), with 96 physical cores and 192 threads.
	
	Both step-wise recurrent inference and sample-wise feedforward inference cover many concrete applications. In this paper, we use text classification to instantiate the step-wise recurrent regime and univariate time-series classification to instantiate the sample-wise feedforward regime.
	
	\subsection{Step-wise recurrent classification}
	
	Text classification is a representative sequential classification task with recurrent processing characteristics, widely adopted benchmarks, and competitive baselines. We therefore evaluate our method on 8 text classification datasets. We report overall accuracy (subset accuracy for multi-label tasks) and inference-time trainable parameter count, and then analyze the results.
	
	\subsubsection{Datasets}
	
	To assess performance comprehensively, we select text classification datasets from Hugging Face, covering short and long texts as well as binary, multi-class, and multi-label settings. For each dataset, we shuffle all samples and split them into 80\% training and 20\% testing sets. We use GoogleNews300d pretrained word vectors and apply 5 epochs of inductive fine-tuning on the training set. The resulting embeddings are then used for both training and testing.
	
	The dataset statistics are summarized in Table~\ref{tab:text_datasets}. 
	
	\begin{table}[htbp]
		\centering
		\caption{Statistics of the text classification datasets used in the step-wise recurrent experiments.}
		\label{tab:text_datasets}
		\begin{tabular}{lrrrr}
			\toprule
			Dataset & Samples & Average Length & Classes & Label \\
			\midrule
			SST-2 & 68221 & 9 & 2 & binary \\
			Rotten Tomatoes & 9596 & 21 & 2 & binary \\
			IMDB & 50000 & 231 & 2 & binary \\
			AG News & 127600 & 37 & 4 & multi-class \\
			Yelp Review Full & 700000 & 134 & 5 & multi-class \\
			DBPedia14 & 630000 & 46 & 14 & multi-class \\
			GoEmotions & 54263 & 12 & 28 & multi-label \\
			Reuters & 10344 & 142 & 67 & multi-label \\
			\bottomrule
		\end{tabular}
	\end{table}
	
	\subsubsection{Settings}
	
	To maximize CPU utilization, the implementation evaluates individuals in parallel using CPU multithreading, with one thread assigned to each individual. As a result, the population size is set to match the number of threads, which is 128 in this work.
	
	For each dataset, we run the proposed method independently for 2500 generations on the training set. To reduce the effect of lucky winners, we do not report the best single individual on the test set. Instead, we select the top 10\% of individuals according to training set accuracy, evaluate them on the test set, and report their mean test accuracy. Each baseline model is trained for up to 50 epochs with early stopping, where patience is set to 5 and the embedding parameters remain frozen. 
	The model settings are summarized in Table~\ref{tab:text_model_settings}. The baseline implementation is based on an open-source repository~\citep{Keras-TextClassification}.
	
	\begin{table}[htbp]
		\centering
		\caption{Model settings for the step-wise recurrent text classification experiments.}
		\label{tab:text_model_settings}
		\begin{tabularx}{\textwidth}{lX}
			\toprule
			Model & Key hyperparameters \\
			\midrule
			FastText & len\_max=56, embed=300, hidden=128, dropout=0.5 \\
			TextCNN & len\_max=50, embed=300, filters=[3,4,5], filters\_num=300, hidden=64, dropout=0.5 \\
			TextRNN & rnn\_type=GRU, num\_rnn\_layers=1, rnn\_units=256, len\_max=50, embed=300, dropout=0.5 \\
			HAN & rnn\_units=256, len\_max=30, len\_max\_sen=50, embed=64, dropout=0.5 \\
			Transformer & encoder\_num=1, head\_num=12, hidden\_dim=3072, len\_max=50, embed=768, dropout=0.1, use\_adapter=False \\
			Seq103 & hidden\_cg\_mutate\_prob=1, cg\_mutate\_prob=1, node\_mutate\_prob=1, conn\_add\_prob=3, conn\_delete\_prob=1.5, hidden\_conn\_add\_prob=3, hidden\_conn\_delete\_prob=1, node\_add\_prob=1, node\_delete\_prob=0.5 \\
			\bottomrule
		\end{tabularx}
	\end{table}

	\subsubsection{Results}
	
	Table~\ref{tab:text_main_results} reports accuracy for binary and multi-class tasks, subset accuracy for multi-label tasks, and inference-time trainable parameter counts. 
	
	\begingroup
	\small
	\setlength{\LTleft}{\fill}
	\setlength{\LTright}{\fill}
	\begin{longtable}{llrrrrr}
		\caption{Main results on the step-wise recurrent text classification benchmarks. ``Params'' denotes the number of inference-time trainable parameters, excluding frozen embeddings. Here, $r^{\mathrm{acc}}=\frac{A_{\mathrm{Seq103}}}{A_{\mathrm{base}}}$ and $r^{\mathrm{comp}}=\frac{P_{\mathrm{base}}}{P_{\mathrm{Seq103}}}$, higher is better for both.}

		\label{tab:text_main_results}\\
		\toprule
		Dataset & Model & Params & Train hours & Accuracy & Compression ratio & Acc ratio \\
		\midrule
		\endfirsthead
		
		\multicolumn{7}{c}{\tablename\ \thetable\ -- continued from previous page} \\
		\toprule
		Dataset & Model & Params & Train hours & Accuracy & Compression ratio & Acc ratio \\
		\midrule
		\endhead
		
		\midrule
		\multicolumn{7}{r}{Continued on next page} \\
		\endfoot
		
		\bottomrule
		\endlastfoot
		
		SST-2 & FastText & 77.2k & 0.002 & 0.810 & 40.6$\times$ & 101.2\% \\
		SST-2 & TextCNN & 1138.7k & 0.154 & 0.917 & 599.3$\times$ & 89.4\% \\
		SST-2 & TextRNN & 461.3k & 0.334 & 0.840 & 242.8$\times$ & 97.6\% \\
		SST-2 & HAN & 1930.8k & 0.738 & 0.914 & 1016.2$\times$ & 89.7\% \\
		SST-2 & Transformer & 1494.4k & 0.418 & 0.849 & 786.5$\times$ & 96.6\% \\
		SST-2 & Seq103 & 1.9k & 31 & 0.820 & -- & -- \\
		\midrule
		Rotten Tomatoes & FastText & 77.2k & 0.001 & 0.705 & 70.2$\times$ & 101.7\% \\
		Rotten Tomatoes & TextCNN & 1138.7k & 0.011 & 0.729 & 1035.2$\times$ & 98.4\% \\
		Rotten Tomatoes & TextRNN & 461.3k & 0.062 & 0.732 & 419.4$\times$ & 98.0\% \\
		Rotten Tomatoes & HAN & 1930.8k & 0.102 & 0.750 & 1755.3$\times$ & 95.6\% \\
		Rotten Tomatoes & Transformer & 1494.4k & 0.039 & 0.502 & 1358.5$\times$ & 142.8\% \\
		Rotten Tomatoes & Seq103 & 1.1k & 8 & 0.717 & -- & -- \\
		\midrule
		IMDB & FastText & 77.2k & 0.002 & 0.817 & 128.7$\times$ & 96.6\% \\
		IMDB & TextCNN & 1138.7k & 0.065 & 0.889 & 1897.8$\times$ & 88.8\% \\
		IMDB & TextRNN & 461.3k & 0.184 & 0.862 & 768.8$\times$ & 91.5\% \\
		IMDB & HAN & 1930.8k & 0.453 & 0.895 & 3218.0$\times$ & 88.2\% \\
		IMDB & Transformer & 1494.4k & 0.363 & 0.836 & 2490.7$\times$ & 94.4\% \\
		IMDB & Seq103 & 0.6k & 86 & 0.789 & -- & -- \\
		\midrule
		AG News & FastText & 77.4k & 0.013 & 0.890 & 70.4$\times$ & 95.4\% \\
		AG News & TextCNN & 1138.8k & 0.138 & 0.912 & 1035.3$\times$ & 93.1\% \\
		AG News & TextRNN & 592.4k & 0.635 & 0.893 & 538.5$\times$ & 95.1\% \\
		AG News & HAN & 2084.4k & 1.604 & 0.918 & 1894.4$\times$ & 92.5\% \\
		AG News & Transformer & 1648.0k & 0.650 & 0.903 & 1498.2$\times$ & 94.0\% \\
		AG News & Seq103 & 1.1k & 33 & 0.849 & -- & -- \\
		\midrule
		Yelp & FastText & 77.6k & 0.025 & 0.543 & 28.7$\times$ & 88.8\% \\
		Yelp & TextCNN & 1138.9k & 1.591 & 0.625 & 421.8$\times$ & 77.1\% \\
		Yelp & TextRNN & 657.9k & 1.760 & 0.562 & 243.6$\times$ & 85.8\% \\
		Yelp & HAN & 2161.2k & 7.668 & 0.643 & 800.4$\times$ & 74.9\% \\
		Yelp & Transformer & 1724.8k & 4.315 & 0.583 & 638.8$\times$ & 82.7\% \\
		Yelp & Seq103 & 2.7k & 442 & 0.482 & -- & -- \\
		\midrule
		DBPedia14 & FastText & 78.7k & 0.018 & 0.968 & 19.2$\times$ & 96.6\% \\
		DBPedia14 & TextCNN & 1139.5k & 1.237 & 0.986 & 278.0$\times$ & 94.8\% \\
		DBPedia14 & TextRNN & 1247.8k & 3.489 & 0.978 & 304.3$\times$ & 95.6\% \\
		DBPedia14 & HAN & 2852.4k & 8.216 & 0.988 & 695.7$\times$ & 94.6\% \\
		DBPedia14 & Transformer & 2416.0k & 2.075 & 0.983 & 589.3$\times$ & 95.1\% \\
		DBPedia14 & Seq103 & 4.1k & 410 & 0.935 & -- & -- \\
		\midrule
		GoEmotions & FastText & 80.5k & 0.007 & 0.233 & 7.3$\times$ & 153.6\% \\
		GoEmotions & TextCNN & 1140.4k & 0.125 & 0.322 & 103.7$\times$ & 111.2\% \\
		GoEmotions & TextRNN & 2165.3k & 0.607 & 0.098 & 196.8$\times$ & 365.3\% \\
		GoEmotions & HAN & 3927.6k & 0.992 & 0.360 & 357.1$\times$ & 99.4\% \\
		GoEmotions & Transformer & 3491.3k & 0.253 & 0.252 & 317.4$\times$ & 142.1\% \\
		GoEmotions & Seq103 & 11.0k & 61 & 0.358 & -- & -- \\
		\midrule
		Reuters & FastText & 85.6k & 0.004 & 0.699 & 2.6$\times$ & 85.6\% \\
		Reuters & TextCNN & 1142.9k & 0.098 & 0.813 & 34.6$\times$ & 73.6\% \\
		Reuters & TextRNN & 4721.2k & 0.084 & 0.539 & 143.1$\times$ & 111.0\% \\
		Reuters & HAN & 6922.8k & 0.189 & 0.755 & 209.8$\times$ & 79.2\% \\
		Reuters & Transformer & 6486.5k & 0.103 & 0.752 & 196.6$\times$ & 79.5\% \\
		Reuters & Seq103 & 33.0k & 42 & 0.598 & -- & -- \\
		
	\end{longtable}
	\endgroup

	Here, the inference-time trainable parameter count is defined as the number of trainable parameters used by the final model at inference time, excluding frozen input embeddings. For the baselines, this corresponds to the non-embedding trainable parameters of the final network. For Seq103, it corresponds to all weights and biases of the evolved final network that are used at inference time, excluding auxiliary quantities used only during exploration, such as innovation numbers and hidden-state archives.
	
	Based on these results, Seq103 exhibits a clear tendency to discover models with far fewer parameters across all datasets while remaining competitive on short-text tasks. On SST-2, the best-performing model is TextCNN, which achieves 0.917 accuracy; Seq103 reaches 0.820 (89.4\% of the best accuracy) with only 1/599.3 of its parameter budget. On Rotten Tomatoes, the best-performing model is HAN (0.750), and Seq103 achieves 0.717 (95.6\%) with only 1/1755.3 of the parameters. Similarly, on DBPedia14, AG News, and GoEmotions, Seq103 achieves 0.935, 0.849, and 0.358 accuracy, corresponding to 94.6\%, 92.5\%, and 99.4\% of the best baseline accuracy, with only 1/695.7, 1/1894.4, and 1/357.1 of the parameters, respectively.
	
	In contrast, accuracy is noticeably lower on IMDB, Yelp, and Reuters. On IMDB, the best model reaches 0.895 (HAN), while Seq103, despite using only 1/3218.0 of its parameters, achieves 0.789 (88.2\% of the best accuracy). On Yelp and Reuters, Seq103 achieves only 74.9\% and 73.6\% of the best-model accuracy, with 1/800.4 and 1/34.6 of the parameters, respectively. Notably, these three datasets all have average text lengths above 100 words (IMDB = 231, Yelp = 134, and Reuters = 142), whereas the others range from 9 to 46. In addition, IMDB is a binary task, which may reduce prediction complexity, so Seq103 still achieves 88.2\% of the best accuracy, but Yelp and Reuters are more challenging because of longer texts together with multi-class or multi-label prediction.
	
	We also observe that Seq103 often achieves performance close to TextRNN across datasets. This is consistent with intuition, since the proposed hidden-state mechanism is inspired by recurrent modeling and thus shows similar behavior to RNN-based methods. However, this similarity may also partly explain why Seq103 can be less competitive than modern attention-based architectures especially on long-text tasks. Compared with models explicitly designed to capture long-range dependencies through attention, the current hidden-state utilization mechanism in Seq103 is still relatively simple.
	
	Collectively, these results suggest that on short-text tasks, Seq103 is capable of delivering over 99\% parameter reduction in 20 out of 25 comparisons, with an average parameter reduction of 98.84\% across all 25 comparisons, while maintaining accuracy close to the best baselines (89.4\% to 99.4\%). However, it still faces challenges on long-text tasks, especially in more complex multi-class and multi-label settings. We believe that improving the modeling of long-range dependencies is a promising direction for future work.

	\subsubsection{Ablation study of step-wise recurrent task}
	
	A natural question is why this method can achieve such a high compression ratio while achieving competitive performance. One possible explanation is that topology itself contributes substantially to representational power, beyond the contribution of continuous parameter values alone: intuitively, topology determines the pathways of information propagation, mathematically, it determines the function composition of the model. It evolves together with other parameters under the constraint of the fitness function. What is ultimately preserved is not only good parameters, but also good structures, they work together so the discovered model can remain effective with a small parameter budget.
	
	To verify this assumption, we designed four ablation experiments by controlling the variability of weights/biases and activation functions, in order to examine the respective contributions of structure, limited parameters, and activation functions to network effectiveness. Unless explicitly mentioned, all other settings are identical to those in Section~4.1, and all experiments are evolved on the SST-2 dataset.
	
	\begin{itemize}
		\item \textbf{Test 1:} All weights and biases are initialized to 1 and then frozen; the activation function is frozen as identity.
		\item \textbf{Test 2:} Based on Test 1, the activation function is included in evolution, while all other settings remain unchanged.
		\item \textbf{Test 3:} Weights and biases are allowed to vary, but their values are restricted to the discrete set [-2, -1, -0.5, 0, 0.5, 1, 2], while the activation function remains frozen as identity.
		\item \textbf{Test 4:} Based on Test 3, the activation function is further included in evolution.
	\end{itemize}
	
	This discrete weight set is designed to represent typical signal transformations: 0 represents elimination, 0.5 represents reduction, 1 represents preservation, 2 represents amplification, and -0.5, -1, and -2 represent inhibition. This configuration is used to examine whether the model can still remain competitive by relying only on topology together with a few basic parameter values. The results are summarized in Table~\ref{tab:ablation_sst2}.
	
	\begin{center}
		\captionof{table}{Ablation study on SST-2.}
		\label{tab:ablation_sst2}
		\begin{tabular}{ccc}
			\toprule
			Test & Acc & Params \\
			\midrule
			Test 1 & 0.6615 & 442 \\
			Test 2 & 0.6661 & 565 \\
			Test 3 & 0.7976 & 284 \\
			Test 4 & 0.8073 & 297 \\
			\bottomrule
		\end{tabular}
	\end{center}
	
	First, by comparing Test 1 with Test 2, and Test 3 with Test 4, we observe that allowing activation functions to evolve does improve model performance, but the improvement is modest. This suggests that diversity of activation functions is beneficial, but not a decisive factor.
	
	Second, by comparing Test 1 with Test 3, and Test 2 with Test 4, we observe that evolving topology alone can achieve performance above chance level, but it is clearly worse than the versions evolved with variable parameters. More importantly, even when the parameters are restricted to a finite discrete set, the model still achieves competitive performance. In Test 4, the model reaches an accuracy of 0.8073, compared with 0.820 in the full evolutionary setting in Section~4.1, corresponding to 98.4\% of the full performance with an additional compression ratio of about 6.39$\times$ (1.9k/297).
	
	These results suggest that competitive performance does not rely solely on large numbers of continuous parameters. Even under severe parameter restrictions, the model can still approach the performance of the full configuration on SST-2. This indicates that expressive power may arise not only from parameter scale, but also from the interaction between topology and parameters.
	
	From a functional perspective, a neural network can be written as $f_T(x;\theta)$, where $T$ denotes the topology and $\theta$ denotes the numerical parameters. Under this view, topology determines the family of functions that the model can represent, while parameters determine a specific instance within that family. Changing the topology therefore changes not only the model structure, but also the underlying function family.
	
	This perspective suggests that the need for many parameters may reflect not only task complexity, but also a mismatch between the chosen function family and the target pattern. In other words, a massive number of parameters may not always reflect the true complexity of the task; some of them may only compensate for a poorly aligned structural prior, that is, an unsuitable function family. In this sense, flexible topology should not be treated as secondary to parameter optimization. Instead, it should be regarded as an important component of model representation, since a better-aligned topology may enable efficient learning with far fewer parameters.

	\subsection{Sample-wise feedforward classification}
	
	Sample-wise feedforward classification does not require the hidden-state mechanism, but it provides an important test for the generality of the Seq103 search strategy. In this setting, hidden connections are disabled, while the other main components, including per-class RMSE-based evaluation, mutation-based evolution with class-wise recombination, and elitism, are retained. Therefore, this experiment evaluates not only whether Seq103 is compatible with feedforward time-series classification, but also whether its modified evolutionary design remains effective beyond the recurrent setting.

	We report overall accuracy and conduct several analyses.
	
	\subsubsection{Datasets}
	
	The full UCR2018 Archive contains 128 univariate datasets across multiple domains. We use the official datasets and their original train--test splits~\citep{UCRArchive2018}. We conduct experiments without z-normalization or any other preprocessing. Shorter sequences are padded with zeros to match the length of the longest sequence within the same dataset. The same input protocol is used for all compared methods.
	
	\subsubsection{Settings}
	
	We run the method on each dataset for 3000 generations, with an early-stopping patience of 300 generations on the training set. Since this experiment does not involve the hidden-state mechanism, hidden connections are not initialized, and the probabilities of adding or deleting hidden connections as well as exchanging hidden-state status are all set to 0. All other settings remain the same as those in Section~4.1. For the baselines, we use open-source implementations containing 9 competitive models~\citep{IsmailFawaz2018deep}, whose configurations are summarized in Table~\ref{tab:ucr_model_settings}.
	
	\begin{table}[tbp]
		\centering
		\caption{Baseline settings for the sample-wise feedforward UCR experiments.}
		\label{tab:ucr_model_settings}
		\begin{tabularx}{\textwidth}{lX}
			\toprule
			Model & Key hyperparameters \\
			\midrule
			ENCODER & conv\_filters=[128,256,512], kernel\_size=[5,11,21], dropout=0.2, attention\_split=256/256 \\
			FCN & conv\_filters=[128,256,128], kernel\_size=[8,5,3] \\
			INCEPTIONTIME & nb\_filters=32, depth=6, use\_residual=True, bottleneck\_size=32, kernel\_size=41, effective=[40,20,10] \\
			MCDCNN & Conv1D(filters=8, kernel\_size=5)$\times$2, MaxPool1D(pool\_size=2)$\times$2 \\
			CNN & conv\_filters=[6,12], kernel\_size=[7,7], avg\_pool\_size=[3,3]\\
			MLP & dense\_units=[500,500,500], dropout=[0.1,0.2,0.2,0.3] \\
			RESNET & n\_feature\_maps=64, residual\_blocks=3, kernel\_size\_per\_block=[8,5,3], channels=[64,128,128] \\
			TLENET & data\_aug: warping\_ratios=[0.5,1,2], slice\_ratio=0.1, conv\_filters=[5,20], kernel\_size=[5,5], pool\_size=[2,4] \\
			TWIESN & alpha(leaky)=0.1, rho$\in[0.55,0.9,2.0,5.0]$, configs=[(N\_x=250, connect=0.5, scaleW\_in=1.0), (250,0.5,2.0), (500,0.1,2.0), (800,0.1,2.0)] \\
			Seq103 & hidden\_cg\_mutate\_prob=0, cg\_mutate\_prob=1, node\_mutate\_prob=1, conn\_add\_prob=3, conn\_delete\_prob=1.5, hidden\_conn\_add\_prob=0, hidden\_conn\_delete\_prob=0, node\_add\_prob=1, node\_delete\_prob=0.5 \\
			\bottomrule
		\end{tabularx}
	\end{table}

	\subsubsection{Results}
	
	Since the complete result table contains more than 1,200 entries, we report the results in summarized form, focusing on classification accuracy, model size (measured by inference-time trainable parameter count), and compression ratio.
	
	For a given baseline model on a given dataset, we define the accuracy ratio and compression ratio as $r^{\mathrm{acc}}=\frac{A_{\mathrm{Seq103}}}{A_{\mathrm{base}}}, r^{\mathrm{comp}}=\frac{P_{\mathrm{base}}}{P_{\mathrm{Seq103}}}.$ Here, $A$ and $P$ denote classification accuracy and parameter count, respectively. For each baseline model, the reported median and mean values are computed over the resulting 128 per-dataset ratios. In addition, Better/Tie/Worse than Seq103 counts the number of datasets on which the baseline performs better than, equal to (difference $< 0.001$), or worse than Seq103, respectively.
	
	\begin{center}
	\captionof{table}{Summary results on the 128 UCR datasets. For each baseline, the reported statistics are computed from the 128 per-dataset values of $r^{\mathrm{comp}}$ and $r^{\mathrm{acc}}$.}

		\label{tab:ucr_summary}
		\resizebox{\textwidth}{!}{
			\begin{tabular}{lccccc}
				\toprule
				Model & Median Compression ratio & Mean Compression ratio ($\pm$std) & Median Acc ratio & Mean Acc ratio ($\pm$std) & Better/Tie/Worse than Seq103 \\
				\midrule
				ResNet & 2084.1$\times$ & 8427.3$\times$ ($\pm$19232.2$\times$) & 88.3\% & 93.0\% ($\pm$55.1\%) & 95/0/33 \\
				FCN & 1095.2$\times$ & 4428.9$\times$ ($\pm$10106.2$\times$) & 90.6\% & 132.1\% ($\pm$442.0\%) & 95/1/32 \\
				CNN & 11.0$\times$ & 36.0$\times$ ($\pm$77.2$\times$) & 99.6\% & 101.4\% ($\pm$20.9\%) & 66/0/62 \\
				Encoder & 13710.3$\times$ & 54538.6$\times$ ($\pm$123096.1$\times$) & 98.8\% & 101.6\% ($\pm$24.3\%) & 68/3/57 \\
				MCDCNN & 2020.8$\times$ & 9095.5$\times$ ($\pm$23636.3$\times$) & 104.5\% & 128.1\% ($\pm$159.3\%) & 42/2/84 \\
				MLP & 2920.4$\times$ & 11591.2$\times$ ($\pm$25318.1$\times$) & 99.9\% & 101.6\% ($\pm$26.2\%) & 63/3/62 \\
				TLENET & 1832.2$\times$ & 7963.8$\times$ ($\pm$20533.8$\times$) & 239.6\% & 396.2\% ($\pm$487.4\%) & 3/0/125 \\
				TWIESN & 1266.4$\times$ & 9575.3$\times$ ($\pm$24624.8$\times$) & 104.0\% & 104.9\% ($\pm$33.0\%) & 52/0/76 \\
				InceptionTime & 1737.8$\times$ & 7026.9$\times$ ($\pm$16036.0$\times$) & 92.1\% & 112.9\% ($\pm$96.8\%) & 83/2/43 \\
				\bottomrule
		\end{tabular}}
	\end{center}
		
	In terms of $r^{\mathrm{acc}}$, Seq103 is closest to CNN, MLP, and Encoder, with median values of 99.6\%, 99.9\%, and 98.8\%, respectively.	The Better/Tie/Worse counts are also nearly balanced for these models, showing that Seq103 performs similarly to them on many datasets. Seq103 performs better than TLeNet and MCDCNN on most datasets. In comparison, ResNet and FCN still achieve higher accuracy on many datasets, and InceptionTime also has an accuracy advantage, although the gap is smaller. Overall, these results show that Seq103 maintains competitive classification accuracy across a large portion of the benchmark.
	
	In terms of $r^{\mathrm{comp}}$, Seq103 uses far fewer parameters than all baseline models, often reducing the parameter count by tens to thousands of times. The median values are consistently high, ranging from 11.0\texttimes{} against CNN to 13,710.3\texttimes{} against Encoder. This shows that Seq103 is generally much more compact across the benchmark. The large gap between the mean and median compression ratios also shows that the compression gain differs greatly across datasets, with some datasets showing much larger reductions than others. Taken together, these results suggest that Seq103 provides a good balance between accuracy and model size, especially when compact models are preferred.

	Figure~\ref{fig:ucr_ratio_curve} shows how Seq103 compares with the baselines on each of the 128 UCR datasets. For each dataset, we compute two dataset-level variants of $r^{\mathrm{acc}}$. The first compares Seq103 with the best baseline on that dataset, and the second compares Seq103 with the median baseline accuracy on that dataset. Specifically, for dataset $d$, let $A_{\mathrm{Seq103},d}$ denote the accuracy of Seq103, let $A^{\mathrm{best}}_d$ denote the highest accuracy among all baselines, and let $A^{\mathrm{med}}_d$ denote the median baseline accuracy. We then define $r^{\mathrm{acc,best}}_d=\frac{A_{\mathrm{Seq103},d}}{A^{\mathrm{best}}_d}, r^{\mathrm{acc,med}}_d=\frac{A_{\mathrm{Seq103},d}}{A^{\mathrm{med}}_d}.$ Each dataset contributes one value to each curve. A value close to 1 indicates that Seq103 achieves accuracy similar to the corresponding reference level, while a value greater than 1 indicates that Seq103 performs better.

	\begin{center}
		\includegraphics[width=0.78\textwidth]{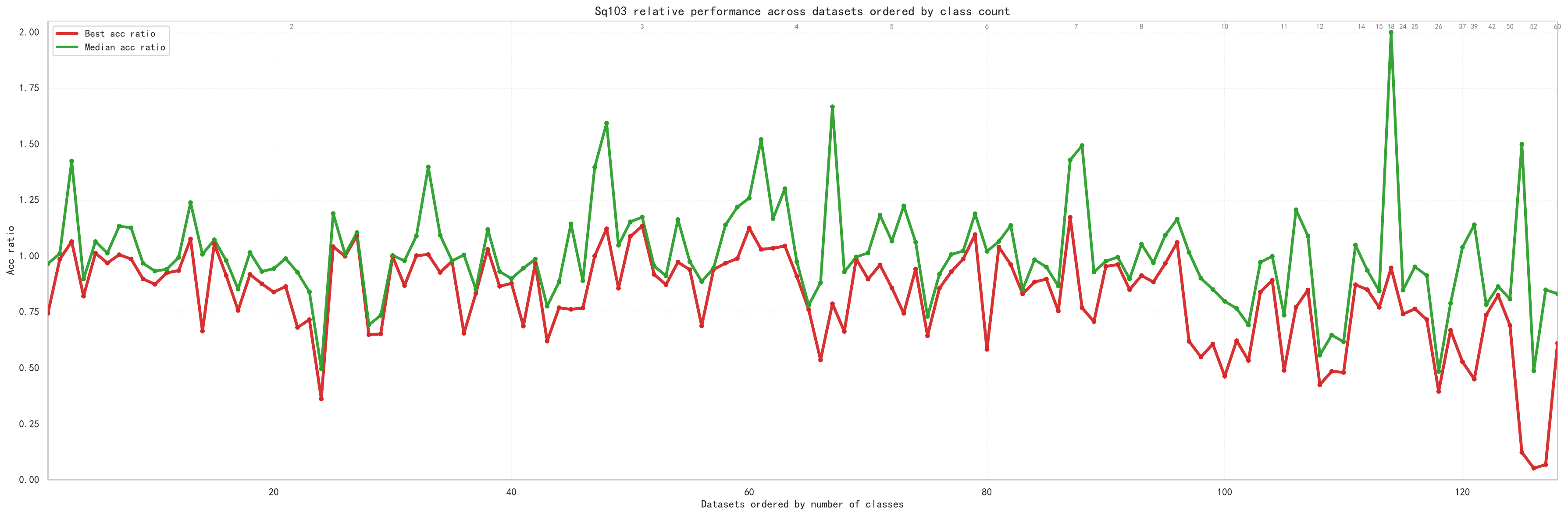}
		\captionof{figure}{Per-dataset accuracy behavior of Seq103 over all 128 UCR datasets.}
		\label{fig:ucr_ratio_curve}
	\end{center}

	As seen in Figure~\ref{fig:ucr_ratio_curve}, $r^{\mathrm{acc,med}}_d$ lies in the range from 0.75 to 1.25 on most datasets, showing that Seq103 generally matches or exceeds the median baseline. However, $r^{\mathrm{acc,best}}_d$ tends to be lower, suggesting that while Seq103 does not always beat the strongest baseline on every dataset, it still holds up well against the overall baseline set. When the datasets are sorted by number of classes, both ratios show a general downward trend with noticeable fluctuations. This suggests that Seq103 becomes less competitive as the class count grows, meaning that tasks with more classes are generally harder for such a compact model. This is consistent with what we observed in Section~4.1 and points to a direction for future improvement.
	
	To further assess the relative standing of Seq103 among competitive baselines, we performed a Friedman test followed by Wilcoxon signed-rank post-hoc comparisons with Holm correction over the 128 UCR datasets. Figure~\ref{fig:cd_diagram} presents the resulting critical difference diagram at significance level $\alpha = 0.05$.
	
	\begin{center}
		\includegraphics[width=0.72\textwidth]{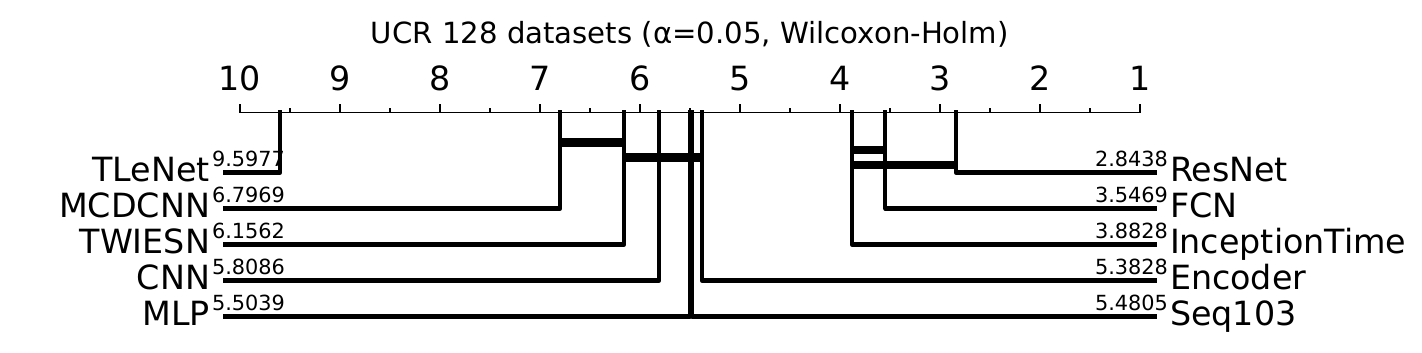}
		\captionof{figure}{Critical difference diagram of Seq103 and the baseline methods on the 128 UCR datasets.}
		\label{fig:cd_diagram}
	\end{center}	

	The Friedman test indicates significant overall differences among the 10 classifiers ($p = 2.47 \times 10^{-92}$). According to the Wilcoxon--Holm post-hoc analysis, Seq103 is not significantly different from CNN, Encoder, MLP, or TWIESN, while ResNet, FCN, and InceptionTime remain significantly stronger, and MCDCNN and TLeNet are significantly weaker. In terms of average rank, Seq103 (5.48) lies close to Encoder (5.38), MLP (5.50), CNN (5.81), and TWIESN (6.16), placing it in a statistically competitive middle group. Considering that Seq103 uses substantially fewer parameters than these models, this result further supports its efficiency--accuracy trade-off.
	
	For further analysis, we select the three datasets with the highest best accuracy ratio and the three datasets with the lowest best accuracy ratio, and present their detailed performance in Table~\ref{tab:ucr_case_studies}.
	
	Table~\ref{tab:ucr_case_studies} reports overall accuracy, inference-time trainable parameter count, $r^{\mathrm{comp}}$, and $r^{\mathrm{acc}}$ relative to each baseline. In the last column, each dataset is summarized by average $r^{\mathrm{comp}}$ / median $r^{\mathrm{acc}}$ / average $r^{\mathrm{acc}}$ / $r^{\mathrm{acc,best}}_d$. Here, the median $r^{\mathrm{acc}}$ is computed over all baselines, the average $r^{\mathrm{acc}}$ excludes the lowest per-baseline value, and $r^{\mathrm{acc,best}}_d$ is computed against the strongest baseline on that dataset.
	
	\begin{center}
	\captionof{table}{Case studies on three best-case and three worst-case UCR datasets for Seq103. The last column reports dataset-level summary statistics in the format of average $r^{\mathrm{comp}}$ / median $r^{\mathrm{acc}}$ / average $r^{\mathrm{acc}}$ / $r^{\mathrm{acc,best}}_d$.}

		\label{tab:ucr_case_studies}
		\resizebox{\textwidth}{!}{
			\begin{tabular}{llrrrrl}
				\toprule
				Dataset & Model & Params & Accuracy & Compression ratio & Acc ratio & Avg compress / median acc / avg acc / best acc \\
				\midrule
				MiddlePhalanxOutlineAgeGroup & FCN & 265.1k & 0.535 & 1244.6$\times$ & 122.5\% & 3015.8$\times$/117.4\%/117.6\%/113.4\% \\
				MiddlePhalanxOutlineAgeGroup & MLP & 543.0k & 0.522 & 2549.3$\times$ & 125.5\% & \\
				MiddlePhalanxOutlineAgeGroup & ResNet & 504.4k & 0.545 & 2368.0$\times$ & 120.1\% & \\
				MiddlePhalanxOutlineAgeGroup & TLENET & 102.6k & 0.571 & 481.5$\times$ & 114.7\% & \\
				MiddlePhalanxOutlineAgeGroup & TWIESN & 643.2k & 0.578 & 3019.7$\times$ & 113.4\% & \\
				MiddlePhalanxOutlineAgeGroup & Encoder & 3.2M & 0.577 & 15018.1$\times$ & 113.7\% & \\
				MiddlePhalanxOutlineAgeGroup & MCDCNN & 102.9k & 0.558 & 482.9$\times$ & 117.4\% & \\
				MiddlePhalanxOutlineAgeGroup & CNN & 783 & 0.534 & 3.7$\times$ & 122.8\% & \\
				MiddlePhalanxOutlineAgeGroup & InceptionTime & 420.6k & 0.565 & 1974.5$\times$ & 116.0\% & \\
				MiddlePhalanxOutlineAgeGroup & Seq103 & 213 & 0.655 & -- & -- & \\
				\midrule
				CinCECGTorso & FCN & 265.2k & 0.829 & 997.1$\times$ & 113.7\% & 4411.1$\times$/125.9\%/172.7\%/112.5\% \\
				CinCECGTorso & MLP & 1.3M & 0.838 & 4973.7$\times$ & 112.5\% & \\
				CinCECGTorso & ResNet & 504.5k & 0.838 & 1896.7$\times$ & 112.5\% & \\
				CinCECGTorso & TLENET & 2.0M & 0.250 & 7680.7$\times$ & 377.0\% & \\
				CinCECGTorso & TWIESN & 10.5k & 0.288 & 39.5$\times$ & 326.8\% & \\
				CinCECGTorso & Encoder & 3.6M & 0.748 & 13542.5$\times$ & 126.1\% & \\
				CinCECGTorso & MCDCNN & 2.4M & 0.801 & 8953.3$\times$ & 117.6\% & \\
				CinCECGTorso & CNN & 9.2k & 0.749 & 34.4$\times$ & 125.9\% & \\
				CinCECGTorso & InceptionTime & 420.7k & 0.272 & 1581.6$\times$ & 346.8\% & \\
				CinCECGTorso & Seq103 & 266 & 0.942 & -- & -- & \\
				\midrule
				InlineSkate & FCN & 265.6k & 0.332 & 349.5$\times$ & 133.4\% & 1816.3$\times$/142.9\%/150.4\%/117.3\% \\
				InlineSkate & MLP & 1.4M & 0.335 & 1902.6$\times$ & 132.2\% & \\
				InlineSkate & ResNet & 504.9k & 0.377 & 664.3$\times$ & 117.3\% & \\
				InlineSkate & TLENET & 2.4M & 0.156 & 3098.1$\times$ & 282.8\% & \\
				InlineSkate & TWIESN & 646.4k & 0.309 & 850.5$\times$ & 142.9\% & \\
				InlineSkate & Encoder & 4.0M & 0.241 & 5297.0$\times$ & 183.2\% & \\
				InlineSkate & MCDCNN & 2.7M & 0.203 & 3606.6$\times$ & 217.6\% & \\
				InlineSkate & CNN & 17.9k & 0.289 & 23.5$\times$ & 153.0\% & \\
				InlineSkate & InceptionTime & 421.1k & 0.358 & 554.1$\times$ & 123.5\% & \\
				InlineSkate & Seq103 & 760 & 0.442 & -- & -- & \\
				\midrule
				PigAirwayPressure & FCN & 271.4k & 0.172 & 67.9$\times$ & 57.0\% & 524.9$\times$/150.0\%/132.5\%/12.3\% \\
				PigAirwayPressure & MLP & 1.5M & 0.065 & 382.4$\times$ & 150.0\% & \\
				PigAirwayPressure & ResNet & 510.7k & 0.406 & 127.8$\times$ & 24.2\% & \\
				PigAirwayPressure & TLENET & 2.5M & 0.019 & 632.6$\times$ & 511.0\% & \\
				PigAirwayPressure & TWIESN & 682.5k & 0.163 & 170.8$\times$ & 60.4\% & \\
				PigAirwayPressure & Encoder & 9.8M & 0.053 & 2463.0$\times$ & 185.5\% & \\
				PigAirwayPressure & MCDCNN & 2.9M & 0.024 & 738.3$\times$ & 408.8\% & \\
				PigAirwayPressure & CNN & 137.3k & 0.061 & 34.4$\times$ & 161.9\% & \\
				PigAirwayPressure & InceptionTime & 426.9k & 0.798 & 106.9$\times$ & 12.3\% & \\
				PigAirwayPressure & Seq103 & 4.0k & 0.098 & -- & -- & \\
				\midrule
				PigArtPressure & FCN & 271.4k & 0.987 & 382.8$\times$ & 5.2\% & 2957.6$\times$/48.7\%/44.3\%/5.2\% \\
				PigArtPressure & MLP & 1.5M & 0.105 & 2154.5$\times$ & 48.7\% & \\
				PigArtPressure & ResNet & 510.7k & 0.991 & 720.3$\times$ & 5.2\% & \\
				PigArtPressure & TLENET & 2.5M & 0.019 & 3564.3$\times$ & 266.0\% & \\
				PigArtPressure & TWIESN & 682.5k & 0.669 & 962.6$\times$ & 7.6\% & \\
				PigArtPressure & Encoder & 9.8M & 0.097 & 13878.1$\times$ & 52.6\% & \\
				PigArtPressure & MCDCNN & 2.9M & 0.029 & 4160.3$\times$ & 177.3\% & \\
				PigArtPressure & CNN & 137.3k & 0.099 & 193.6$\times$ & 51.6\% & \\
				PigArtPressure & InceptionTime & 426.9k & 0.875 & 602.1$\times$ & 5.8\% & \\
				PigArtPressure & Seq103 & 709 & 0.051 & -- & -- & \\
				\midrule
				PigCVP & FCN & 271.4k & 0.831 & 71.7$\times$ & 7.8\% & 553.7$\times$/84.9\%/80.4\%/6.8\% \\
				PigCVP & MLP & 1.5M & 0.076 & 403.4$\times$ & 84.9\% & \\
				PigCVP & ResNet & 510.7k & 0.918 & 134.9$\times$ & 7.0\% & \\
				PigCVP & TLENET & 2.5M & 0.019 & 667.3$\times$ & 336.0\% & \\
				PigCVP & TWIESN & 682.5k & 0.519 & 180.2$\times$ & 12.4\% & \\
				PigCVP & Encoder & 9.8M & 0.066 & 2598.2$\times$ & 97.3\% & \\
				PigCVP & MCDCNN & 2.9M & 0.015 & 778.9$\times$ & 418.9\% & \\
				PigCVP & CNN & 137.3k & 0.071 & 36.2$\times$ & 90.6\% & \\
				PigCVP & InceptionTime & 426.9k & 0.952 & 112.7$\times$ & 6.8\% & \\
				PigCVP & Seq103 & 3.8k & 0.065 & -- & -- & \\
				\bottomrule
		\end{tabular}}
	\end{center}
	
	As shown in Table~\ref{tab:ucr_case_studies}, Seq103 achieves its best results on InlineSkate, MiddlePhalanxOutlineAgeGroup, and CinCECGTorso. These datasets have a moderate number of classes, with 7, 3, and 4 classes, respectively. This suggests that Seq103 remains competitive when the task structure is relatively simple and can be handled by a compact model.
	
	By contrast, Seq103 performs worst on the Pig-series datasets, including PigAirwayPressure, PigArtPressure, and PigCVP. These datasets belong to the same 52-class physiological signal setting. They are complex because they have many classes and more complex signal patterns, which usually require greater model capacity. This performance gap suggests that Seq103 is more suitable for simpler time-series classification tasks with fewer classes, especially in settings where model size is a key constraint.
	
	These results show that the modified framework can discover very lightweight sequence classifiers. This makes the resulting models useful for deployment settings where storage or computational resources are limited. However, this benefit comes at a cost, as the search process is computationally expensive.
	
	Part of this cost likely comes from the current implementation. Unlike the baseline models, which are implemented in PyTorch or TensorFlow, our system runs on a different framework and may miss some low-level optimizations provided by mainstream deep learning libraries. As a result, a direct comparison of inference latency would not be fully fair, because the measured latency could reflect framework differences rather than only model efficiency. Therefore, we do not report latency comparisons in this work. Instead, we report the training time only as a rough reference for computational cost.

	\subsubsection{Ablation study of sample-wise feedforward task}
	
	To better understand the source of the sample-wise UCR results, we compare Seq103 with two original NEAT variants under the same sample-wise setting. This comparison has two goals. First, we want to check whether the accuracy gain comes from the Seq103 modifications or only from the basic NEAT pipeline. Second, we want to understand whether the small model size mainly comes from the NEAT-style search space.
	
	Since original NEAT is not a standard baseline for time-series classification, there is no commonly used fitness function for time-series classification tasks, so we design two variants. The first variant uses the same per-class RMSE-based fitness as Seq103. The second variant uses overall accuracy as the fitness.
	
	The comparison with the per-class RMSE-based variant helps us check whether the advantage of Seq103 only comes from the fitness design. This baseline uses the same fitness signal as Seq103, but it does not include the other Seq103 modifications. Therefore, if Seq103 performs better, the gain is more likely to come from the full Seq103 framework, including individual recombination, per-class RMSE, and elitism.
	
	The accuracy-fitness variant helps rule out the possibility that the loss-fitness baseline performs poorly simply because original NEAT is not suitable for a loss-based fitness such as per-class RMSE. Since overall accuracy is a simple and intuitive fitness choice, it serves as a second baseline for the original NEAT pipeline.
	
	\begin{table}[t]
		\centering
		\small
		\caption{Comparison between Seq103 and two original NEAT variants on the 128 UCR datasets. The median accuracy ratio is reported as $r^{\mathrm{acc}}=\frac{A_{\mathrm{Seq103}}}{A_{\mathrm{base}}}$. To show that Seq103 typically uses more parameters than the original NEAT variants, we additionally report the median size ratio $r^{\mathrm{size}}=\frac{P_{\mathrm{Seq103}}}{P_{\mathrm{base}}}$. B/T/W counts the number of datasets on which the baseline performs better than, equal to, or worse than Seq103, respectively. A tie means that the accuracy difference is smaller than 0.001.}
		
		\label{tab:seq103_vs_original_neat}
		\begin{tabularx}{\linewidth}{>{\raggedright\arraybackslash}Xccccc}
			\toprule
			Baseline & B/T/W & Med. Acc diff & Med. Acc ratio & Med. Size ratio & Wilcoxon p \\
			\midrule
			Loss-fitness NEAT & 17/2/109 & 0.104 & 1.199 & 11.675 & 5.78e-17 \\
			Accuracy-fitness NEAT & 37/3/88 & 0.056 & 1.114 & 13.026 & 7.46e-9 \\
			\bottomrule
		\end{tabularx}
	\end{table}
	
	Table~\ref{tab:seq103_vs_original_neat} shows that Seq103 clearly improves accuracy over both original NEAT variants. For the loss-fitness NEAT baseline, the Better/Tie/Worse count is 17/2/109, meaning that Seq103 performs better on 109 datasets. For the accuracy-fitness NEAT baseline, the count is 37/3/88, meaning that Seq103 performs better on 88 datasets. The median values of $r^{\mathrm{acc}}$ are 1.199 and 1.114, respectively. The Wilcoxon signed-rank tests also show that the accuracy gains are statistically significant, with p-values of 5.78e-17 and 7.46e-9.
	
	These results suggest that the performance gain does not come only from using the basic NEAT pipeline. Since Seq103 outperforms the variant with the same per-class RMSE fitness, the improvement is more likely due to the coupled design choices of Seq103 as a unified framework.
	
	The parameter results show a different pattern. Seq103 usually uses more parameters than these two variants, with median $r^{\mathrm{size}}$ values of 11.675 and 13.026. This is expected because Seq103 adds extra mechanisms to improve accuracy. However, the absolute number of parameters is still very small. This suggests that the lightweight nature of Seq103 is largely inherited from the NEAT-style search space, which builds networks from simple nodes and connections and can naturally produce compact models.
	
	Among the two variant baselines, the accuracy-fitness version performs slightly better than the loss-fitness version. However, we do not use overall accuracy as the fitness in Seq103 because overall accuracy gives only one score for the whole individual. It does not provide class-level feedback. In contrast, the class-level individual recombination mechanism in Seq103 needs class-level evaluation signals, which are provided by the per-class RMSE design.
	
	It should also be noted that individual recombination and per-class RMSE are tightly linked in Seq103. The recombination mechanism depends on the per-class evaluation signal, so these two components cannot be fully separated in a clean ablation study. For this reason, we treat them as part of one unified framework and interpret the observed improvement as the result of the overall Seq103 design.
	
	Overall, this ablation study shows that Seq103 improves the accuracy of the original NEAT pipeline in the sample-wise setting. At the same time, its small model size is mainly inherited from the NEAT-style search space, which favors compact networks made of simple nodes and connections. Although Seq103 uses more parameters than the two original NEAT variants, it remains very small in absolute terms and achieves much better accuracy. This suggests that Seq103 keeps the lightweight advantage of NEAT while improving its performance through the proposed modifications.
	
	\subsection{Further Discussion: Beyond Classification}
	\label{sec:beyond_classifier}
	
	In this paper, Seq103 provides a set of elementary computational ingredients for exploring how adaptive systems may be built through evolution, and we test these ingredients on sequence classification tasks. These ingredients include nodes, connections, hidden-state units for temporal memory, and an evolutionary search process that can modify both parameters and topology. Beyond compact classification, we believe that such a framework suggests several promising research directions, especially toward adaptive intelligent systems.
	
	This view is motivated by a simple observation: although the principles of intelligence and consciousness are still far from being fully understood, evolution is the only known process that has produced natural intelligence. Therefore, we think it is meaningful to study whether artificial systems can also acquire useful structures, memory mechanisms, and adaptive behaviors through evolutionary search. In this context, Seq103 can be viewed as a small step in that direction. It does not only search numerical parameters, but also changes the structure of the network itself. More importantly, the hidden-state mechanism gives the evolved structure a basic ability to preserve and reuse information over time, which is necessary for long-term interaction.
	
	A natural future direction is to connect this type of evolutionary system with real-world environments. In the current experiments, feedback comes from supervised classification tasks. In a more general setting, feedback could come from continuous interaction with an environment. Multimodal signals could be used as inputs, and the outputs of the evolved network could be mapped to actions or control signals for software agents, robots, or embodied systems. Under such a setting, the system would no longer be only a classifier, but a compact adaptive structure that evolves under environmental feedback.
	
	A key conceptual issue behind this direction is the definition of evolution itself. As a foundation for this line of research, it is necessary to clarify whether the process implemented by current evolutionary algorithms should be regarded as genuine evolution or only as a simulation of evolution. In other words, observable phenomena such as increasing structural complexity, expanding populations, or improving fitness may not be sufficient to represent the key mechanisms of evolution in the natural world. We cannot resolve this question in this paper, but we believe that judging artificial evolution only by surface-level observations is not rigorous enough.
	
	This also leads to a broader question: if it remains controversial whether a system that simulates language behavior possesses true intelligence, then why should a system that simulates evolutionary behavior be treated as an exception? To avoid another form of ``stochastic parrot'', future research should develop a more rigorous definition of artificial evolution, together with clearer criteria for evaluating whether an artificial evolutionary process captures the essential mechanisms of natural evolution. In this sense, defining evolution is not only a philosophical issue, but also a necessary technical foundation for building adaptive systems through evolutionary search.
	
	\section{Conclusion}
	
	This paper presents Seq103, a unified neuroevolution framework for sequence classification. By extending a shared elementary search space with optional hidden-state mechanisms, Seq103 supports both sample-wise feedforward classification and recurrent sequence modeling under the same framework, reducing reliance on predefined architectures across different processing regimes.

	The discovered models are highly parameter-efficient while remaining competitive with strong baselines. On step-wise tasks, Seq103 achieves 73.6\% to 99.4\% of the accuracy of the best baselines while reducing parameters by 34.6$\times$ to 3218.0$\times$, with an average accuracy retention of 86.96\% and an average compression ratio of 1120.35$\times$. On sample-wise tasks, Seq103 achieves 5.2\% to 117.3\% of the accuracy of the best baselines while reducing parameters by 11.8$\times$ to 160,601.0$\times$, with an average accuracy retention of 81.95\% and an average compression ratio of 9312.07$\times$.
	
	These results show that flexible topology search can provide a practical route to compact sequence classifiers across different data regimes. At the same time, Seq103 still faces clear limitations on long-text tasks and datasets with large label spaces, where stronger long-range dependency modeling and greater representational capacity may be required. It should also be noted that the search process remains computationally expensive, making Seq103 more suitable for scenarios in which architecture search can be performed offline in advance and the resulting compact model is then deployed for inference. Improving search efficiency, enhancing temporal dependency modeling, and developing more scalable class-wise recombination mechanisms remain important directions for future work. Beyond classification, the framework also suggests a possible direction for studying compact adaptive systems built through evolutionary search.

	\printcredits
	\section*{Declaration of generative AI and AI-assisted technologies in the manuscript preparation process}
	
	During the preparation of this work, we used ChatGPT by OpenAI to improve grammar and language readability. After using this tool, the authors reviewed and edited the manuscript as needed and assume full responsibility for the content of the published article.

	\bibliographystyle{cas-model2-names}
	\bibliography{references}
	
\end{document}